\documentclass{article}
\PassOptionsToPackage{numbers, compress}{natbib}

\usepackage[preprint]{neurips_2024}

\usepackage{hyperref}
\hypersetup{
colorlinks=true,
linkcolor=black,
citecolor=black
}
\usepackage{url}            
\usepackage{booktabs}       
\usepackage{amsfonts}       
\usepackage{nicefrac}       
\usepackage{microtype}      
\usepackage{xcolor}         
\usepackage{tabularx, makecell, multirow,graphicx}
\usepackage{amsmath}
\usepackage{lastpage}
\usepackage{booktabs}  
\usepackage{fancyhdr}
\usepackage{subfigure}
\usepackage{makecell}
\usepackage{graphicx} 
\usepackage{indentfirst}
\usepackage{cellspace}

\title{Continuous Approximations for Improving Quantization Aware Training of LLMs}

\author{
  He Li \\
  Beijing Academy \\
  \texttt{li@imlihe.com} \\
  \And
  Jianhang Hong \\
  Wuhan Britain-China School \\
  \texttt{hongjianhang30@gmail.com} \\
  \And
  Yuanzhuo Wu \\
  Luwan Senior Highschool \\
  \texttt{yuanzhuo\_wu@outlook.com} \\
  \AND
  Snehal Adbol  \\
  Ryan International School \\
  \texttt{thesnehaladbol@gmail.com} \\
  \And
  Zonglin Li \\
  Chongqing Bashu Secondary School \\
  \texttt{mccreelee@outlook.com}
 }

\begin{document}
\maketitle

\begin{abstract}
  Model compression methods are used to reduce the computation and energy requirements for Large Language Models (LLMs). Quantization Aware Training (QAT), an effective model compression method, is proposed to reduce performance degradation after quantization. To further minimize this degradation, we introduce two continuous approximations to the QAT process on the rounding function, traditionally approximated by the Straight-Through Estimator (STE), and the clamping function. By applying both methods, the perplexity (PPL) on the WikiText-v2 dataset of the quantized model reaches 9.0815, outperforming 9.9621 by the baseline. Also, we achieve a 2.76\% improvement on BoolQ, and a 5.47\% improvement on MMLU, proving that the step sizes and weights can be learned more accurately with our approach. Our method achieves better performance with the same precision, model size, and training setup, contributing to the development of more energy-efficient LLMs technology that aligns with global sustainability goals.
\end{abstract}

\section{Introduction}

Large Language Models (LLMs), due to their high computational power and memory requirements, have extremely high energy consumption in the training and inference phases \cite{stojkovic2024towards}. 
Model compression methods are used to reduce the energy and computation requirements of LLMs \cite{tang2024survey}.

Quantization is a model compression method that reduces computational complexity by lowering the precision of model weights and activations.
Many quantization techniques for Transformers have been proposed in recent years \cite{tang2024survey}, including many effective Post Training Quantization (PTQ) methods such as SmoothQuant \cite{xiao2023smoothquant}, OmniQuant \cite{shao2023omniquant}, and APTQ \cite{guan2024aptq}. Quantization Aware Training (QAT) is an effective technique for model compression \cite{nagel2021white} that often outperforms PTQ methods \cite{tang2024survey}.

Although there are not many related works dedicated to QAT of LLMs, the LLM-QAT work has achieved remarkable results and provided a good foundation for further studies \cite{liu2023llm}. This study proposes  approaches for QAT of Transformers, including QAT for key-value cache and data-free QAT through knowledge distillation.

We aim to improve the accuracy and stability of the quantization process by enabling more effective learning of step size and quantized model weights. To enhance QAT in LLMs training, we adopt Sigmoid STE and propose SoftClamp, which are continuous approximations of the rounding function and clamping functions, respectively. We mainly delve into the theoretical underpinnings of our proposed methods and the empirical results demonstrating their effectiveness. 
\section{Background}
\textbf{Custom estimators} 
To simulate quantization losses, QAT introduces a rounding function, the derivative of which at all differentiable points is 0. To generate non-zero gradients, STE was introduced \cite{bengio2013estimating}, which is widely used in scenarios when the gradient of neuron network modules is not applicable for back-propagation \cite{yin2019understanding}. For example, when training Binarized Neural Networks \cite{NIPS2016_d8330f85}, STE is applied to approximate the originally non-differentiable binary networks. Considering the function $f(x) = \mathrm{round}(x)$, STE approximates $\frac{df}{dx}$ as $1$. Therefore, the gradient is passed directly during the back-propagation process.

Many research attempts to improve the STE. A kind of research has improved the performance by designing new estimators, mostly differentiable approximation functions of STE, including PWL \cite{PWL}, MAD \cite{MAD}, HGTE \cite{HTGE}, and so on. These works that propose customized estimators usually refer to the concept of ``gradient error'': the gradient descent process is negatively affected by the fact that STE does not fit well to the rounding function \cite{HTGE, SIGMOID}. However, a recent work suggests that gradient error does not affect performance under certain conditions \cite{schoenbauer2024custom}.

\textbf{Selection of Scaling Factor} In QAT using standard STE with symmetric quantization, the quantization loss is formulated by following equations:
\begin{equation} \label{QAT}
Q(x) = s \times \mathrm{round}\left[ \mathrm{clamp} (\frac{x}{s}, a, b)  \right],  
\hspace{1cm}
\mathrm{clamp}(x, a, b) = \max[ \min(x, b), a ],
\end{equation}
where $s$ is the scaling factor, which can be selected in many ways. For example, MinMax \cite{jacob2018quantization, krishnamoorthi2018quantizing}, also known as dynamic quantization, determines $s$ by the range of actual inputs at each level. For symmetric quantization, the value of $s$ by MinMax is $
s = \frac{ \max( \textbf{X} ) }{ 2^{N-1} - 1 },
$ where $\textbf{X}$ is the input tensor, and $N$ is the target quantization bit width. Another work introduces a learnable $s$: by applying STE, the gradient of $s$ is available, which allows $s$ to be a learnable parameter \cite{esser2019learned}.
\section{Methods}
Based on \autoref{QAT}, we adopt Sigmoid STE \cite{SIGMOID} and propose SoftClamp. The two methods are applied to the following two mechanisms respectively:

\textbf{Rounding:} \autoref{QAT} involves rounding the clamped value. This simulates the precision limitation in quantization. 

\textbf{Clamping and Scaling: }
    The clamping function $\mathrm{clamp}(x, a, b) = \max[ \min(x, b), a ] $ in QAT ensures that the input values are constrained within a specified range $[a,b]$, simulating the range constraint of quantized representations. 
\subsection{Sigmoid STE}
Sigmoid STE is a continuous approximation of the rounding function that replaces the STE, and experimental results support performance improvements by Sigmoid STE \cite{SIGMOID}. 

\begin{figure}[ht]
    \centering
    \subfigure[Floor and sigmoid estimators. \label{sub:left}]{\includegraphics[width=0.45\linewidth]{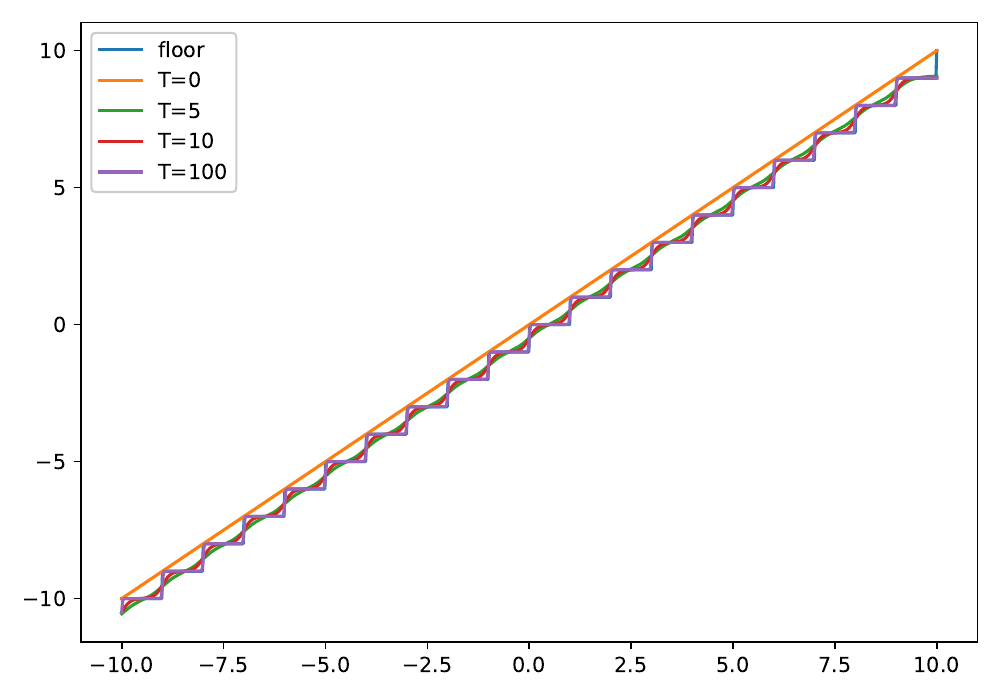}}
    \hspace{1em} 
    \subfigure[Sigmoid estimator derivatives. \label{sub:right}]{\includegraphics[width=0.45\linewidth]{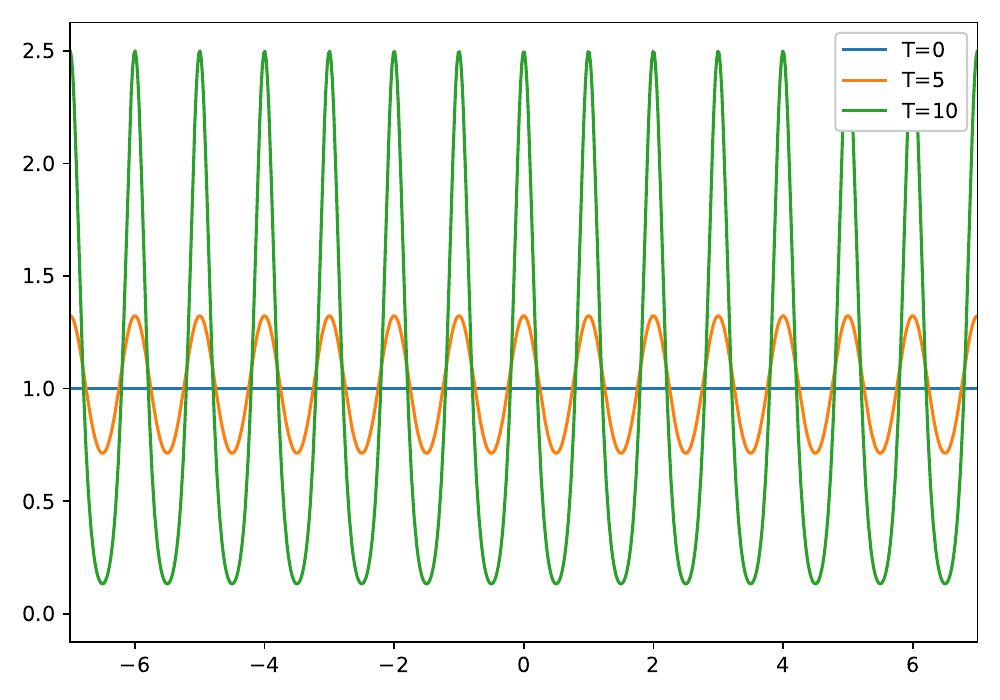}} 
    \caption{Sigmoid STEs and their derivatives. }
    \label{fig:sigmoid}
\end{figure}

Similar to other custom estimators, Sigmoid STE is used to mitigate the gradient error, which is considered to have a negative impact on model's performance \cite{SIGMOID, HTGE}. The Sigmoid STE replaces the rounding function during backpropagation \autoref{QAT}. The following expressions show the formulation of Sigmoid STE, $f$, and its derivative, $\frac{df}{dx}$:
\begin{equation}
    f(x,T) = \sum_{i=0}^{ \left \lfloor x \right \rfloor  } \frac{1}{1 + \exp(T (x - i) )},
    \hspace{2em}
    \frac{df}{dx} = \sum_{i=0}^{ \left \lfloor x \right \rfloor } T \cdot \frac{\exp(T(x - i))}{(1 + \exp(T(x - i)))^2}.
\end{equation}
As shown in Figure \ref{sub:left}, the parameter \( T \) in Sigmoid STE controls the degree of approximation. A higher value of \( T \) leads to sharper transitions , allowing a more accurate approximation of the rounding function. This increased sensitivity also modulates the gradient, as shown in Figure \ref{sub:right}.
\subsection{SoftClamp}
\begin{figure}[ht]
    \centering
    \subfigure[$\hat{C}$ with $a=-10$ and $b=10$. \label{fig:SoftClamp}]{\includegraphics[width=0.48\linewidth]{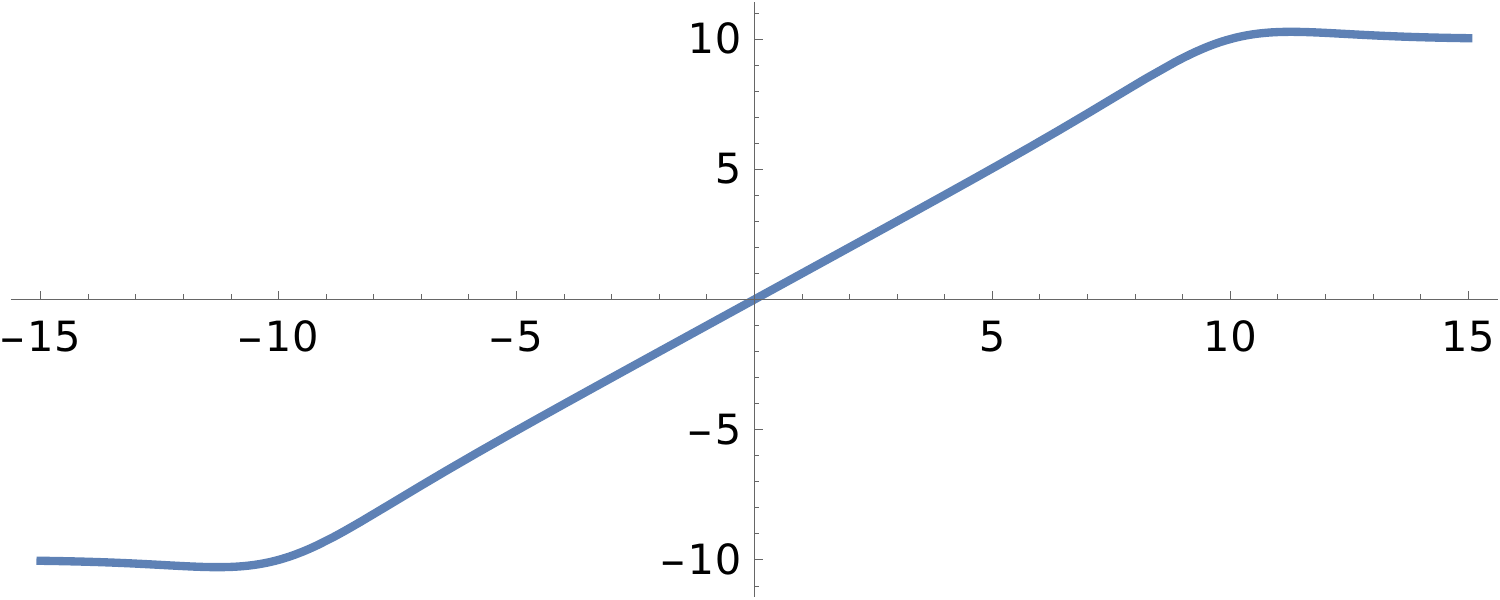}} 
    \hspace{1em}
    \subfigure[$C$ with $a=-10$ and $b=10$. \label{fig:clamp}]{\includegraphics[width=0.48\linewidth]{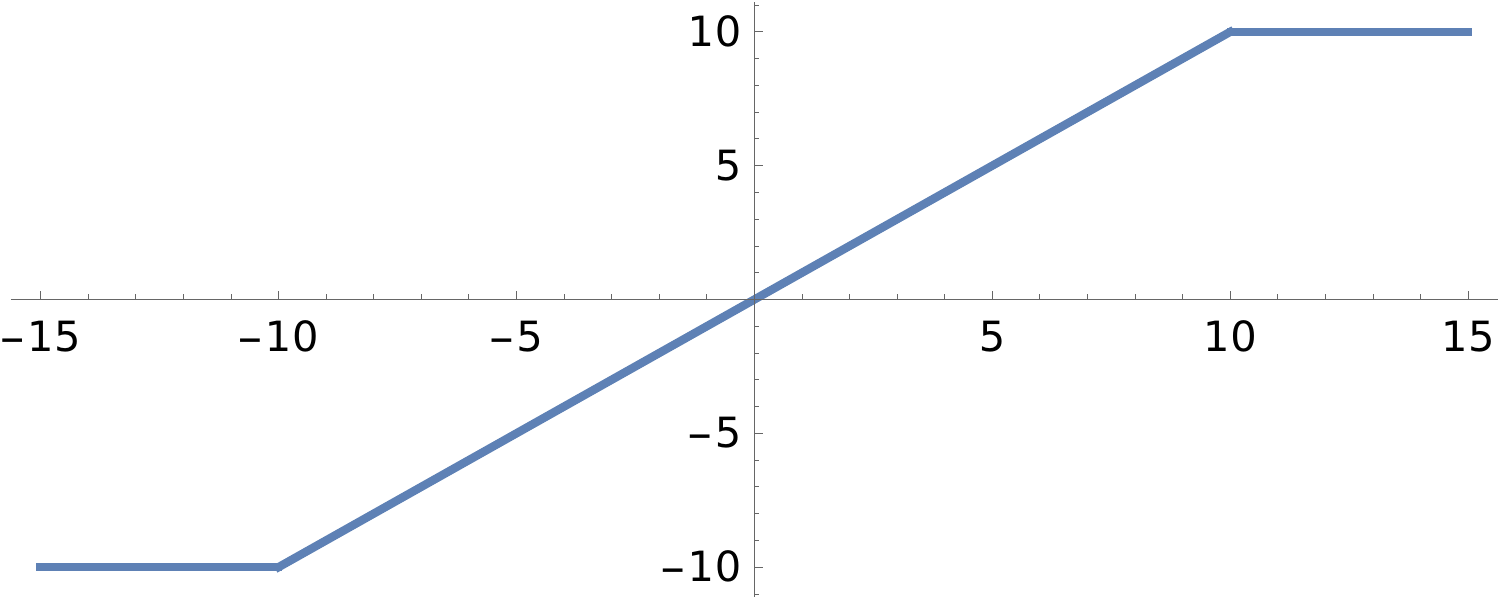}}
    \vspace{0em}
    \subfigure[Derivative of $\hat{C}$ with respect to $s$ with $x=20$. \label{fig:Sds}]{\includegraphics[width=0.48\linewidth]{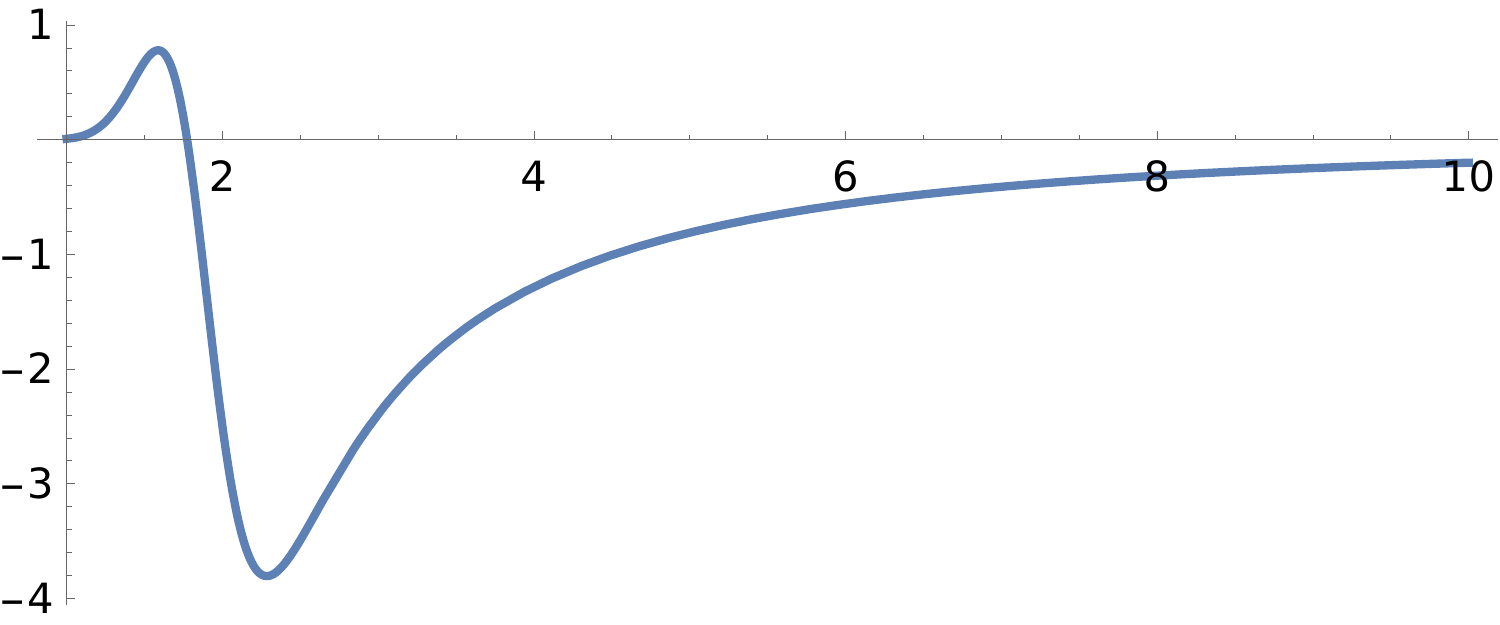}} 
    \hspace{1em} 
    \subfigure[Derivative of $C$ with respect to $s$ when $x=20$. \label{fig:ds}]{\includegraphics[width=0.48\linewidth]{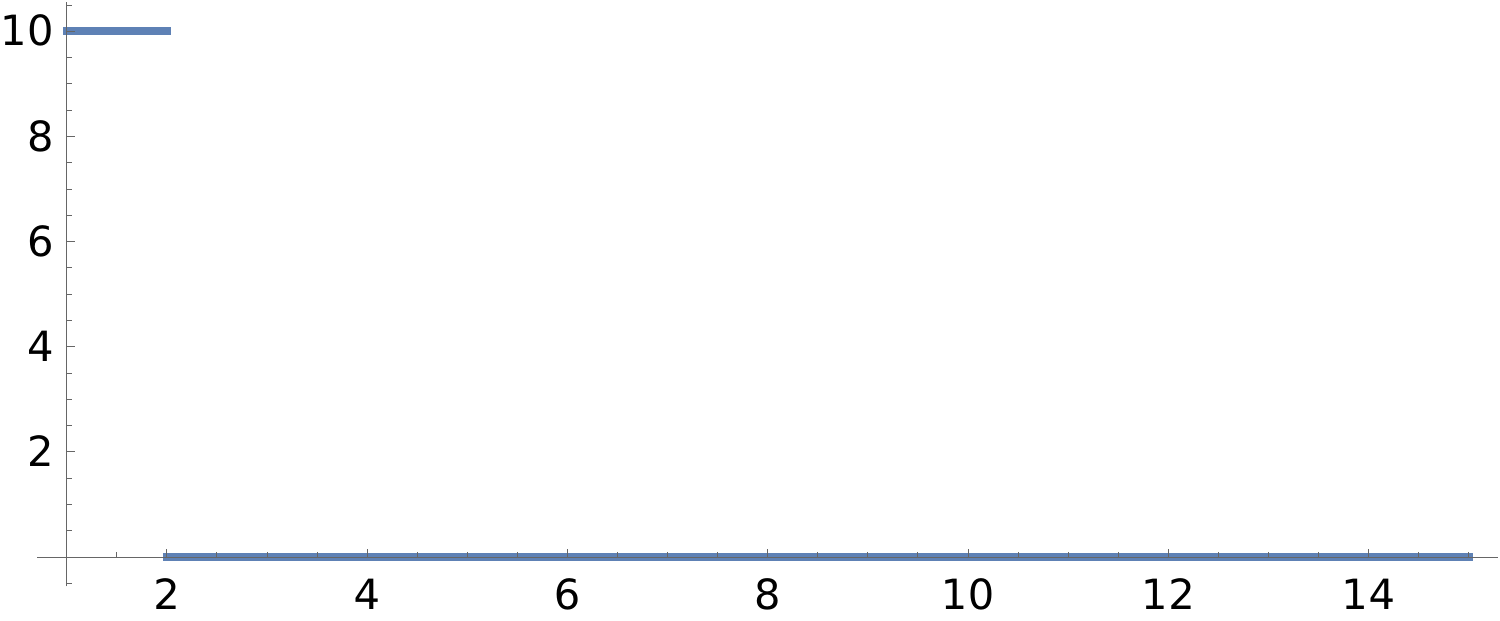}}

    \caption{Plots of SoftClamp function and derivatives.}\label{fig:soft}
\end{figure}

Recently, nonlinear rectifier units achieved great success in LLMs \cite{shazeer2020glu}. Inspired by this, we propose the SoftClamp mechanism, which can be formulated as
\begin{equation}
\mathrm{SoftClamp}(x, a, b) := x \times g(x - a) \times g(b - x) + a \times g(a - x) + b \times g(x - b),
\end{equation}
where $a$ and $b$ are the lower and upper limits, respectively. $g$ is a continuous function that introduces non-linearity into the network, taking the form of any S-shaped functions such as those used in Gaussian Error Linear Unit (GeLU) \cite{hendrycks2016gaussian}, Sigmoid Linear Unit (SiLU) \cite{elfwing2018sigmoid}, or Swish GLU (SwiGLU) \cite{ramachandran2017searching}. In our implementation and figures below, we use SwiGLU with $\beta = 1$ following LLaMA-2 and LLaMA-3 \cite{touvron2023llama}. Then, we define the part of \autoref{QAT} that involves clamping as the function $C$ and involves SoftClamp $\hat{C}$:
\begin{equation}
    \hat{C}(x) = \mathrm{SoftClamp} \left ( \frac{x}{s}, a, b \right ),
    \hspace{2em} C(x) = \mathrm{clamp} \left ( \frac{x}{s}, a, b \right ).
\end{equation}
Figure \ref{fig:SoftClamp} shows the plot of $y=\hat{C}(x, -10, 10)$, Figure \ref{fig:clamp} shows that for $C$, Figure \ref{fig:Sds} shows the plot of $\frac{d \hat{C}}{d s}$ when $x=20$, and Figure \ref{fig:ds} shows the plot of $\frac{d C}{d s}$ with the same $x$ value. From the derivative plots, it can be seen that under the conventional method when the value of $s$ allows $x$ to be represented within the accuracy limits ($s \ge 2$), the gradient of $s$ will vanish. Instead, our method generates a gradient of $s$ that is non-zero, allowing it to be updated. $s$ is responsible for balancing representation range and accuracy \cite{esser2019learned}.  We believe that without vanishing gradient, $s$ can be learned better, improving the overall performance.

\section{Experiments}
\subsection{Experimental Setup}

We evaluate performance for both generation and question answering.  For the generation part, we fine-tuned   LLaMA-3-8B \cite{llama3modelcard} on a subset of the FineWeb \cite{huggingfacefw_2024}. The evaluating metric is perplexity (PPL) on the WikiText-v2-test dataset \cite{merity2016pointer}. For question answering part, we fine-tuned the instruct version of LLaMA-3-8B on auxiliary training questions in MMLU \cite{hendryckstest2021}. Then, models are evaluated on BoolQ \cite{clark2019boolq} and MMLU test sets. We use 4-bit precision for weights and KV cache values, and 8-bit precision for activations.


\subsection{Abalation}
Based on our methods, models are fine-tuned separately, each with just vanilla STE ($T=0$), vanilla STE and SoftClamp ($T=0$ + SC), or Sigmoid STE and SoftClamp ($T>0$ + SC) applied. 
\begin{table}[ht]
\caption{PPL on WikiText-v2-test} \label{tab:ppl}
\begin{center}
    \vspace{-1mm}

    \begin{tabular}{l|llll}
        \toprule
         Methods & $T=0$ & $T=5$ & $T=10$ & $T=100$      \\
        
        \hline

        Without SoftClamp & 9.9621 & 9.9152  & 9.8997 & 10.0523 \\  
        With SoftClamp & 9.7194 & 9.5814 & 9.2792 & \textbf{9.0815} \\
        
        \bottomrule
    \end{tabular}
\end{center}
\end{table}

According to \autoref{tab:ppl}, while applying SoftClamp independently does improve the metric, no significant improvements are spotted when Sigmoid STE is applied to the model alone, which is consistent with the work that challenged the significance of custom estimators \cite{schoenbauer2024custom}. 
 However, when both methods are applied, PPL decreases significantly. As the $T$, which controls the degree of approximation, grows, PPL decreases significantly from 9.7194 ($T=0$) to 9.0815 ($T=100$). We believe that when combining two methods, the step size $s$ and model weights can be adapted more efficiently to the quantization loss during the QAT process, thereby improving the performance.

\subsection{Question Answering}

We also tested our method on the question answering dataset. The results on BoolQ and MMLU datasets are shown in \autoref{tab:qa}.

\begin{table}[ht]
\caption{Scores on Q\&A datasets.} \label{tab:qa}
\begin{center}
    \vspace{-2mm}

    \begin{tabular}{l|llll}
        \toprule
         Dataset & $T=0$ & $T=100$ & $T=0 + \mathrm{SC}$ & $T=100 + \mathrm{SC}$ \\
        
        \hline

        BoolQ & 71.1848 & 70.9420  & 72.7049 & \textbf{73.1483} \\  
        MMLU & 59.1463 & 59.8057 & 60.1491 & \textbf{62.3829} \\
        
        \bottomrule
    \end{tabular}
\end{center}
\end{table}

The results of both metrics show the capacity of our method, demonstrating improvements of  2.76\% on BoolQ, and 5.47\% on MMLU.

\section{Conclusion}
In conclusion, we introduce continuous approximations of the rounding and clamping functions to improve QAT for LLMs. The Sigmoid STE and SoftClamp methods, when applied together, lead to better performance. We believe the continuous approximations create more sound gradient for both step size and weights, obtaining better parameters. Therefore, causing less performance degradation, QAT will be a more promising method for creating LLMs that require less computational power and energy.

While our study demonstrates significant advancements, there are limitations to consider. Due to time and hardware limits, the dataset we used was relatively small, and we did not explore other combinations of precision, potentially limiting the scope of the improvements. Future research should address these areas to further optimize QAT for LLMs.
\newpage
\bibliography{refs}
\appendix
\section{Appendix}
\subsection{Experimental Details}
All models are trained with a single epoch. Our models are trained with five RTX4090s

During all evaluation tasks, the temperature is set to zero.

For the FineWeb dataset, we utilize the CC-MAIN-2024-18 subset \cite{huggingfacefw_2024}, which consists of 154.4 billion tokens.

BoolQ \cite{clark2019boolq}: Evaluated in a zero-shot setting using a User/Assistant style prompt. The score is reported as the percentage of correct responses.      

MMLU \cite{hendryckstest2021}: The score is calculated as the average of the subscores across different tasks. A 5-shot approach is used, with examples selected randomly from questions with the same topic.
\end{document}